\documentclass[10pt,journal,compsoc]{IEEEtran}
\ifCLASSOPTIONcompsoc
  \usepackage[nocompress]{cite}
\else
  \usepackage{cite}
\fi
\usepackage{cite}
\usepackage{graphicx}
\usepackage{multicol}
\usepackage{subfig}
\usepackage{amsmath}
\usepackage{amsfonts}
\usepackage{romannum}
\usepackage{caption}
\usepackage{tabularx}
\usepackage{tabulary}
\usepackage{array}
\usepackage{tabu}
\usepackage{verbatim}
\usepackage{multirow}
\usepackage{makecell}
\usepackage{longtable}
\usepackage{adjustbox}

\usepackage{textcomp}
\usepackage{xcolor}
\usepackage{makecell}

\usepackage{dblfloatfix}

\ifCLASSINFOpdf
\else
\fi
\begin{document}
\title{Robust Kernel-based Feature Representation for \\3D Point Cloud Analysis via \\Circular  Convolutional Network}
\author{Seunghwan Jung, Yeong-Gil Shin, and Minyoung Chung$^{\ast}$%
\thanks{\textit{Asterisk indicates corresponding author.}}%
\thanks{\textit{(This paper is under consideration at Computer Vision and Image Understanding.)}}%
\thanks{S. Jung and Y.-G. Shin are with the Department of Computer Science and Engineering, Seoul National University, South Korea.}
\thanks{*M. Chung is with the School of Software, Soongsil University, South Korea (e-mail: chungmy@ssu.ac.kr).}}

\IEEEtitleabstractindextext{%
\begin{abstract}
Feature descriptors of point clouds are used in several applications, such as registration and part segmentation of 3D point clouds. Learning representations of local geometric features is unquestionably the most important task for accurate point cloud analyses. However, it is challenging to develop rotation or scale-invariant descriptors. Most previous studies have either ignored rotations or empirically studied optimal scale parameters, which hinders the applicability of the methods for real-world datasets. In this paper, we present a new local feature description method that is robust to rotation and scale variations. Moreover, we improved representations based on a global aggregation method. First, we place kernels aligned around each point in the normal direction. To avoid the sign problem of the normal vector, we use a symmetric kernel point distribution in the tangential plane. From each kernel point, we first projected the points from the spatial space to the feature space, which is robust to multiple scales and rotation, based on angles and distances. Subsequently, we perform convolutions by considering local kernel point structures and long-range global context, obtained by a global aggregation method. We experimented with our proposed descriptors on benchmark datasets (i.e., ModelNet40 and ShapeNetPart) to evaluate the performance of registration, classification, and part segmentation on 3D point clouds. Our method showed superior performances when compared to the state-of-the-art methods by reducing 70$\%$ of the rotation and translation errors in the registration task. Our method also showed comparable performance in the classification and part-segmentation tasks without any external data augmentation.
\end{abstract}

}

\maketitle

\IEEEdisplaynontitleabstractindextext

% \fi
\IEEEpeerreviewmaketitle

\IEEEraisesectionheading{\section{Introduction}\label{sec:introduction}}

\IEEEPARstart{P}{oint} cloud analysis is becoming a popular research area owing to the growth in the capability of 3D sensors to capture rich geometric 3D information. The applications of point cloud analysis include robotics, autonomous driving, and augmented/mixed reality. Extracting salient local geometric information is a fundamental task for analyzing point clouds to match correspondences between two objects \cite{SmoothNet} or to analyze the geometric information \cite{DGCNN}. Recently, end-to-end learning based on point or graph convolutional networks has outperformed earlier works, which were primarily developed using hand-crafted feature descriptors \cite{SHOT}\cite{FPFH}. However, building rotation- or scale-invariant descriptors remains a difficult task in the field of computer vision research.

% ================================================ %
% Paragraph 2. 관련 연구 brief + common limitiation
Descriptors of point cloud applications have been widely researched for point cloud registration, model segmentation, and classification. PointNet \cite{PointNet} shows a new paradigm for point cloud analysis by introducing a permutation-invariant method; however, it is difficult to encode the local geometric information. The point pair feature network (PPFNet) \cite{PPFNet} encodes local features by employing PointNet \cite{PointNet} for local regions and a deep graph convolutional neural network (DGCNN) \cite{DGCNN} encodes the relative position of neighbors for each point. However, these methods are limited to extracting rotation-invariant features. {\color{black} In practice, it must be noted that the point cloud is not aligned to the same frame, indicating that random rotation of the input point cloud can significantly affect the representation of the descriptors.} Kernel point convolution (KPConv) \cite{KPConv} uses kernel points around each point to efficiently handle irregularly distributed point clouds. KPConv has demonstrated groundbreaking performance; however, this rotation-variant descriptor limits the performance for randomly rotated objects, which are obtained by multiview scans. 3DSmoothNet \cite{SmoothNet} extracts local region points and aligns the local points to the local reference frame of the center point. The primary limitation of 3DSmoothNet is that the sign of a normal axis and the directions of the other two axes are not unique in a planar region. Descriptors that are aligned by an inaccurate local reference frame may encode different geometric contexts. In the example of point cloud registration, if the corresponding points have different normal signs, descriptors from the points may hamper the identification of correspondences. Moreover, local descriptors only encode local geometric information, which results in difficulty in encoding the global geometry. Consequently, local descriptors of monotonous and repeating areas are typically considered to be nonsalient descriptors, which indicates that global registration can be mismatched.

% ================================================ %
% Paragraph 3. To overcome the previous limitation
To overcome this limitation, we propose a rotation- and {\color{black}scale-robust} descriptor-generation method. Inspired by KPConv \cite{KPConv} and 3DSmoothNet \cite{SmoothNet}, our proposed method aligns the kernels to the normal vector and extracts rotation-robust features. Owing to the nonuniqueness of the local reference frame in the planar region, we distributed kernels in the form of a cylindrical shape. This shape is symmetrical around a tangent plane to handle the sign problem and has a circular cross section to handle the other undefined reference axes. By employing this kernel structure, we applied convolution with adjacent kernels combined together such that the descriptor is not affected by rotation. To make the descriptor robust to the scale of the local frame, we analyzed the geometric information and rebuilt the descriptor with a modified kernel size. In addition, to improve representations of the descriptor from the monotonous and repeating areas, we aggregated all features based on the distances from each point to encode discriminative global features.

The major contributions of this work can be summarized as follows:
\begin{itemize}
    \item The rotation-robust descriptor is developed based on the kernel alignment.
    \item The sign problem, which is caused by the normal direction of the vector, is resolved by the proposed angle-based convolution.
    \item The scale factor, which is derived from the size of a kernel, is automatically defined for each point using a scale adaptation module.
    \item Global context is effectively extracted by the proposed aggregation method with each local context.
\end{itemize}
We analyzed the working of our proposed method on three types of tasks: registration, classification, and part segmentation. We trained and tested our proposed method on ModelNet40 dataset \cite{ShapeNet} for classification and registration and on ShapeNet dataset \cite{ShapeNetPart} for segmentation.
\par

The remainder of this paper is organized as follows. In Section II, several hand-crafted and deep-learning-based 3D features are reviewed. The proposed method is described in Section III. The experimental results, discussion, and conclusion are presented in Sections IV, V, and VI, respectively.

\section{Related Works}
\subsection{Hand-crafted 3D features}
Before the advance of deep learning, a 3D feature descriptor was developed based on hand-crafted methods. Local descriptors were generated based on the relationship between a point and the spatial neighborhoods around the point. In addition, certain methods built a rotation-invariant descriptor based on a local reference frame. Spin-images \cite{SPIN} align neighbors using the surface normal of the interest point and represent aligned neighbors to the cylindrical support region using radial and elevation coordinates. The 3D Shape Context descriptor \cite{3DShapeContext} represents neighbors in the support region with grid bins divided along the azimuth, elevation, and radial values. The Unique Shape Context method \cite{USC} extends the 3D Shape Context method by applying a local reference frame based on the covariance matrix. Similarly, the signature of histograms of orientations algorithm \cite{SHOT} also calculates the local reference frame and builds a histogram using angles between point normal vectors. Point feature histograms \cite{PFH} and fast point feature histograms \cite{FPFH} select neighbors for each point and build a histogram using pairwise geometric differences between neighbors and the point of interest, such as relative distance and angles. Recently, with the advent of deep neural networks for point cloud data (e.g., PointNet \cite{PointNet} and DGCNN \cite{DGCNN}), feature descriptors have shown groundbreaking results when compared to hand-crafted methods in several vision tasks.

\subsection{Deep learning based 3D features}

\subsubsection{Volumetric based Methods}
The conversion of the point cloud to a volumetric data representation has been widely used to employ grid-based convolutions \cite{Voxel01, Voxel02}. However, the quantification of the floating-point data results in an approximation, such that the input data intrinsically contains discretized artifacts. Because the voxelization process severely consumes memory, these methods typically approximate the input data into a coarse grid of volumetric representation. To overcome this problem, certain methods represent point cloud data by optimizing the memory consumption. OctNet \cite{Voxel04Octree} divides the space by employing a set of unbalanced octrees based on density. Certain methods use a sparse tensor that only saves the nonempty space coordinates and features \cite{Sparse, Minkowski, FCGF}.
\par
To build a rotation-invariant descriptor, 3DsmoothNet \cite{SmoothNet} was used to calculate the local reference frame based on the covariance of points and to transform neighbor points within the spherical support area of the interest point using the local reference frame before voxelizing the points. However, the sign of a normal axis and the directions of the other two axes are not unique in the planar region. Descriptors that are aligned using an inaccurate local reference frame may encode different geometric contexts. Therefore, we assume that the sign of the normal vector is not unique and uses a customized kernel similar to the KPConv \cite{KPConv} method to overcome the sign issue.
{\color{black} SpinNet \cite{spinnet} aligned each point and neighbors of the point (i.e., patch) with the z-axis and mapped the point patch to the cylindrical volume to build rotation-invariant descriptors. However, since the method used the volumetric-based method for each point patch, it required a lot of computational memory to build descriptors. Moreover, since each volume contained only one point patch, the user has to determine the optimal patch size to be trained (i.e., fixed scale).
On the contrary, our proposed method consumed a relatively small amount of computational memory when compared to SpinNet \cite{spinnet} and automatically determined the kernel size using the scale adaptation module.
}

\begin{figure}[t]
\begin{center}
\includegraphics[width=1.0\linewidth]{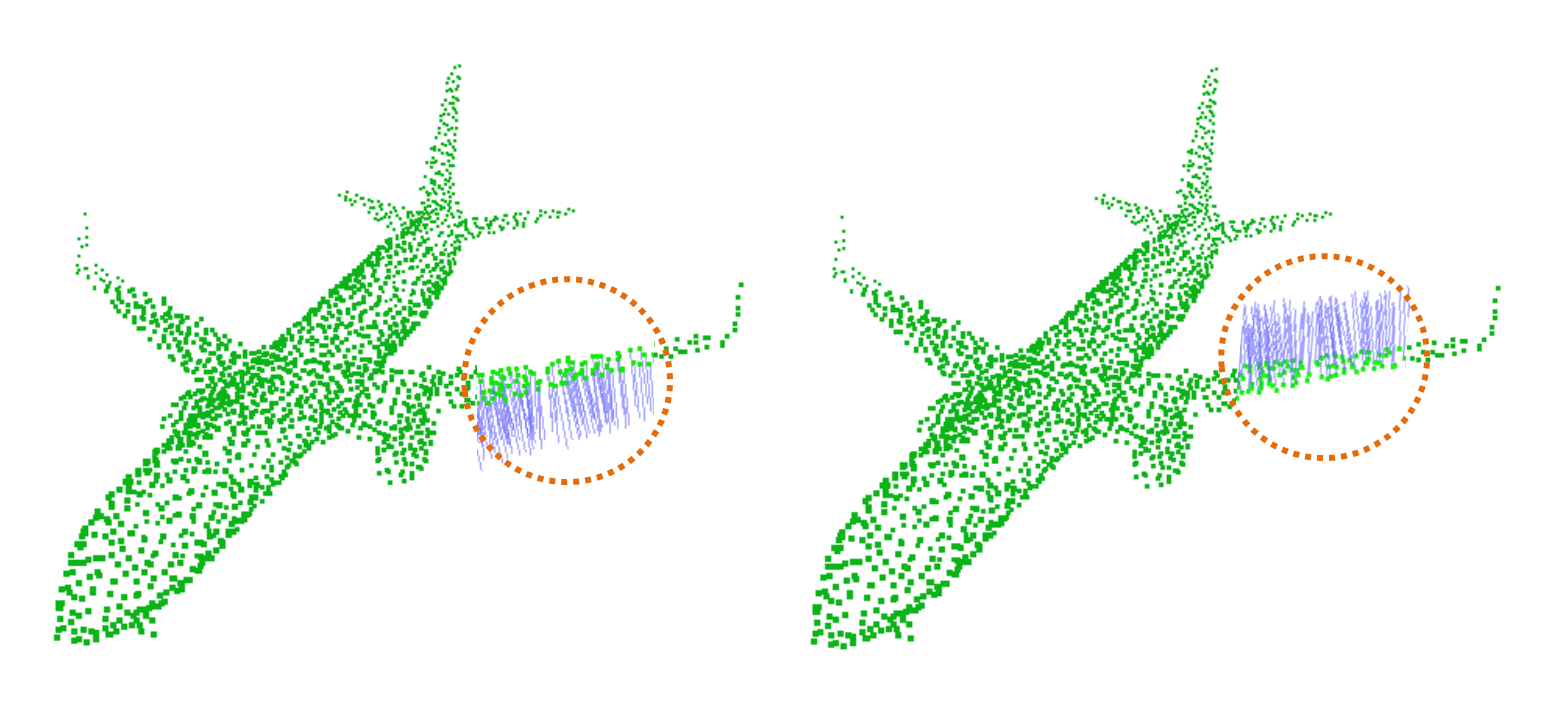}
\end{center}
  \caption{Visualization of normal vectors. The signs for each point on a planar surface are not determined uniquely.}
\label{fig:sign}
\end{figure}

\subsubsection{Point based Methods}
PointNet \cite{PointNet} and PointNet++ \cite{PointNet++} are pioneering works for point cloud analysis, which are based on deep neural networks. These methods encode unstructured point clouds using a shared multilayer perceptron and build a permutation-invariant descriptor using a global max-pooling layer. Based on PointNet, various methods have been developed to improve performance. PPFNet \cite{PPFNet} extended PointNet \cite{PointNet} to learn local geometric features. PPFNet built local features by employing PointNet and subsequently fused global information based on the local features by employing max-pooling. PPF-FoldNet \cite{PPF-FoldNet} used rotation-invariant features such as angles and distances between the interest point and its neighbor and trained the descriptor by using folding-based auto-encoding in an unsupervised fashion. DGCNN \cite{DGCNN} selected k-nearest neighbors for each point and encoded the relative locations of the neighbors to encapsulate the geometric information. 
{\color{black} ShellNet \cite{ShellNet} partitioned the neighbors of each point into shells based on the distances from the point to resolve the point order ambiguity.}
KPConv \cite{KPConv} proposed a kernel-based point convolution method that placed kernel points around each point to effectively handle irregularly distributed point clouds, and further, aggregated the geometric information from the kernel points.
We extended the KPConv method by employing normal kernel alignment and angle-based convolution. As described in KPConv, a normal vector is available for artificial data \cite{KPConv}. In the real-world dataset, a local reference frame is inaccurate because of the sign problem (i.e., the uncertain direction of the normal vector; Fig. \ref{fig:sign}). To overcome the inaccuracy of the local reference frame, we aligned the kernels and extracted features based on the unsigned normal axis, and subsequently applied convolution operations that are invariant to the sign problem.
{\color{black} Various rotation-invariant methods have been studied based on the rotation-invariant features, such as relative distance, angle \cite{RI-ShellConv}\cite{Clusternet}, and quaternion \cite{REQNN}.
RIF \cite{TVCG2021} represented neighbors of the interest point by using rotation-invariant features and constructed a point relation matrix to supplement insufficient global information. The method showed better performances when compared to the other methods under rotations, but showed inferior performances under non-rotation environments when compared to non-rotation-invariant methods.
It is challenging to develop a rotation-robust descriptor with a good benchmark accuracy because of two reasons: 1) It is hard to represent the accurate relationship between points with rotation-invariant features. 2) Convolution with more than one point in rotation-invariant-order is a challenging problem. In that aspect, our proposed method represented neighbors with not only the rotation-invariant features, but the kernels to supplement the relationship representation. Further, we used the circular convolution method, which processed the adjacent kernels simultaneously to capture the relationships between the points.
}

\begin{figure}[t]
\begin{center}
\includegraphics[width=1.0\linewidth]{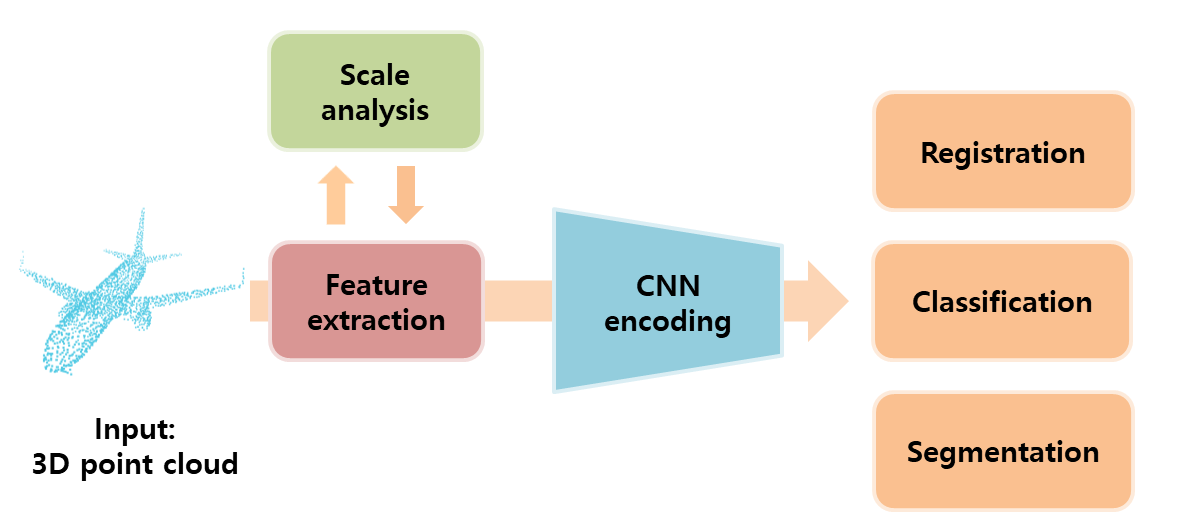}
\end{center}
   \caption{Overview of the proposed architecture. First, features are extracted using multiple kernel sizes. Subsequently, scale analysis is employed based on the interpolation between the kernel sizes. Finally, the feature descriptor is encoded using the adjusted scale for the downstream tasks.}
\label{fig:overview}
\end{figure}

\begin{figure*}[t]
\begin{center}
\includegraphics[width=0.8\linewidth]{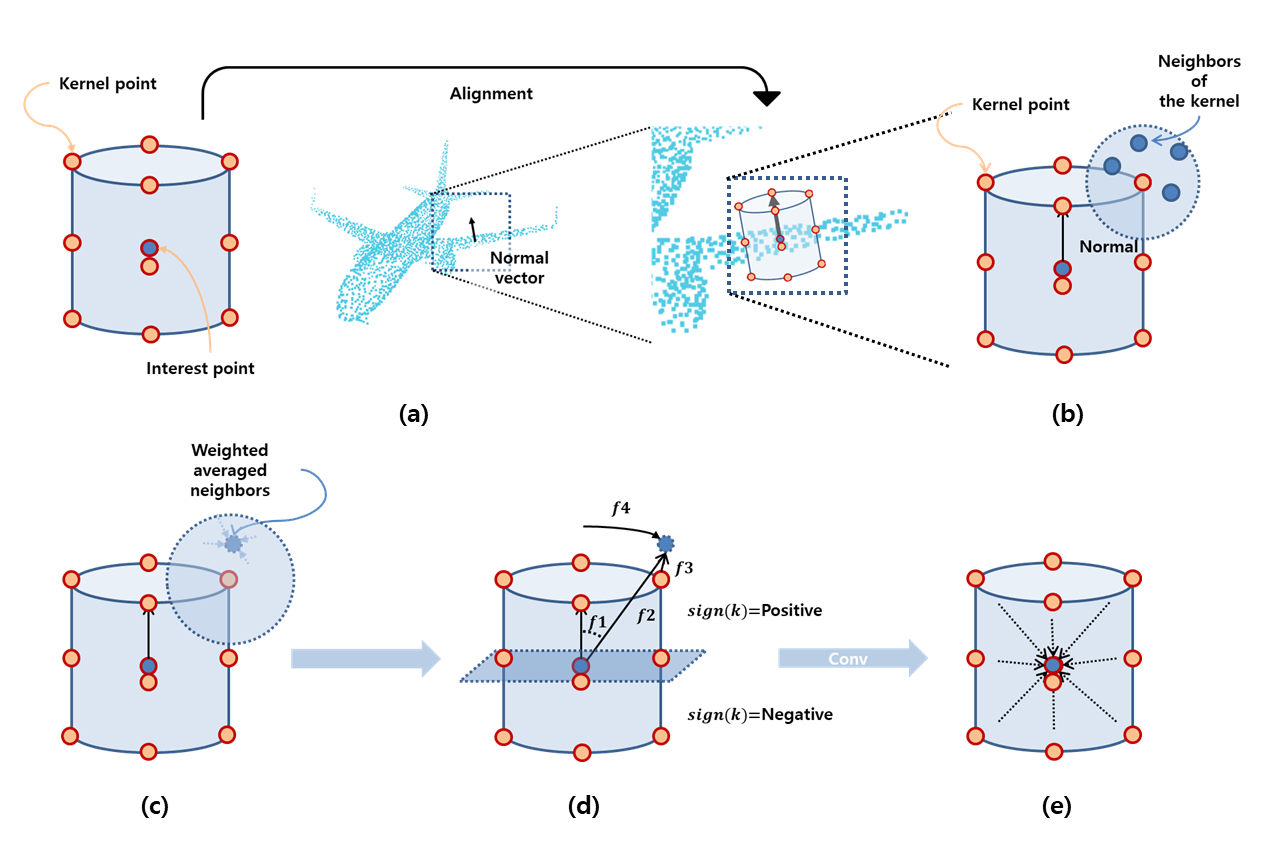}
\end{center}
   \caption{(a) Kernel points are aligned using the normal vector of the target point; (b) Neighbors are selected for each kernel point; (c) Weighted-average location is estimated based on the distance from the kernel point to the neighbors; (d) For each kernel, the relative location of the averaged neighbor is estimated using distances and angles; (e) After convolution, kernel features are aggregated by summation and maximum value selection to represent the interest point.}
\label{fig:kernel_01}
\end{figure*}

\begin{figure*}[t]
\begin{center}
\includegraphics[width=0.8\linewidth]{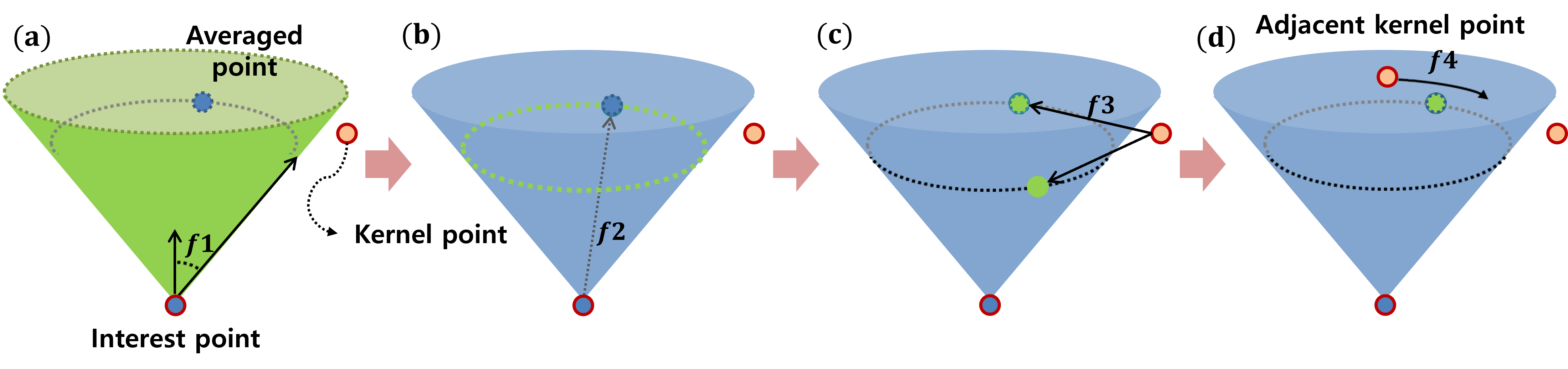}
\end{center}
   \caption{{\color{black}Candidates of the neighbor points corresponding to extracted features. Green regions of (a), (b), (c), and (d) illustrate the candidates corresponding to the features (f1), (f1, f2), (f1, f2, f3), and (f1, f2, f3, f4), respectively.}}
\label{fig:candidates}
\end{figure*}

\begin{figure}[t]
\begin{center}
\includegraphics[width=1.0\linewidth]{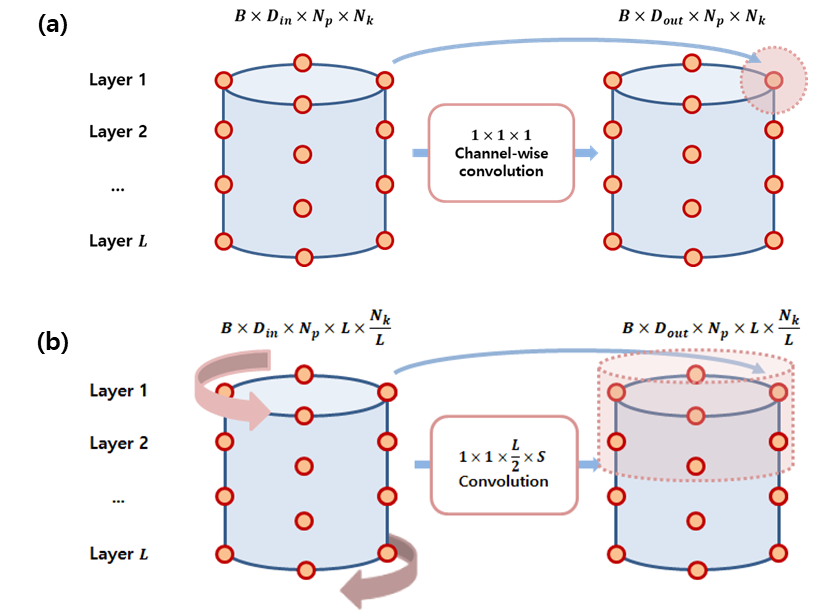}
\end{center}
   \caption{(a) Channel-wise convolution and (b) circular convolution. The red transparent region indicates the receptive field. S indicates the kernel size of the convolution.}
\label{fig:kernel_03}
\end{figure}

\section{Method}
Figure \ref{fig:overview} illustrates an overview of the proposed descriptor-generation framework. The point descriptor is built by using information obtained from the kernels around the point. To build rotation-invariant kernels, we use the normal vectors of each point to align the kernels (Section A). Local information is extracted from the kernels to encode the feature descriptors (Section B). Subsequently, circular convolution is applied, which is invariant to the sign problem (Section C). A scale adaptation module is employed in the network to resolve the scale issues (Section D). Various convolutional neural network (CNN) encoder architectures are presented in Section E, which were used for downstream tasks in this study. Finally, the global context estimation is demonstrated, which is employed in the encoder architectures (Section F).

\subsection{Kernel alignment}
To build a rotation-robust descriptor, we aligned the kernels around each point using the local reference axis, i.e., normal vector. 
{\color{black} Inspired by the 3DsmoothNet \cite{SmoothNet}, we estimated the normal vector using the eigenvector of the neighbor point covariance matrix.}
However, the sign of the normal vector and the remaining local reference axes are not unique if a point is located on a planar surface. To resolve these ambiguities, we used cylinder-shaped kernels in which the cylinder column is aligned to the normal vector.

Figure \ref{fig:kernel_01}(a) illustrates the kernel distribution. The cross section of the cylinder is a circle along the normal direction. We placed the kernels for each circle (i.e., four to six numbers of kernels), and grouped them as one layer. In total, we used three layers for the cylinder (i.e., additional upper and lower regions of the tangent plane).

\subsection{Rotation robust feature projection}
Once all the kernels are aligned for each point, the k-nearest neighbor points from each kernel are extracted. The averaged location is then calculated based on their distance from the kernel:
\begin{eqnarray}
\hat{x}^k_i = \sum_{x_j \in N(x^k_i)} \frac{w_j x_j}{\sum w_j} \text{    where    } w_j = exp(\frac{-(x_j - x^k_i)^2}{d^2}),
\label{eq:eq_kernel_01}
\end{eqnarray}
where $\hat{x}^k_i \in \mathbb{R}^{Nx3}$ is the weighted-average location of the $k$-th kernel point of $x_i$ and $d$ indicates the distance from the center point to the kernel point. For the weighting term, i.e., $w_j$ we used the Gaussian function to reduce the influence of outliers in (\ref{eq:eq_kernel_01}).
\par

For rotation-robust representations, we estimated four types of features. We first estimated the angles between two vectors: one is from the center point of the kernels to the weighted-average point and the other is the normal vector (the angle f1 in Fig. \ref{fig:kernel_01}(d)). However, because the normal vector has a normal orientation problem (i.e., sign ambiguity), a negative sign is multiplied with the normal vector if the kernel is located below the tangent plane, as shown below: 
\begin{eqnarray}
%{f1}^k_i = |v_i \cdot \frac{\hat{x}^k_i - x_i}{||\hat{x}^k_i - x_i||}|
{f1}^k_i = v_i \cdot \frac{\hat{x}^k_i - x_i}{||\hat{x}^k_i - x_i||} \cdot sign(k)
\label{eq:eq_feature_01}
\end{eqnarray}
where $v_i$ indicates the normal vector of $x_i$ and $sign(k)$ returns a negative sign if the kernel is located below the tangent plane. This term determines the angle value, regardless of the normal sign.

Next, we estimated the distances from the point to the averaged neighbors and from the center of the kernels to the averaged neighbors (distances f2 and f3 in Fig. \ref{fig:kernel_01}(d)): 
\begin{eqnarray}
{f2}^k_i  = ||\frac{\hat{x}^k_i - x_i}{d}||
\label{eq:eq_feature_02}
\end{eqnarray}
\begin{eqnarray}
{f3}^k_i  = ||\frac{\hat{x}^k_i - x^k_i}{d}||
\label{eq:eq_feature_03}
\end{eqnarray}

\noindent
To provide direction to the closest adjacent kernel points, we estimated the distance ratio from two adjacent kernel points to the averaged point (ratio f4 in Fig. \ref{fig:kernel_01}(d)): 
\begin{eqnarray}
%{f4}^k_i = w_{k-1}{f*}^{k-1}_i + w_{k+1}{f*}^{k+1}_i
{f4}^k_i = \frac{||\hat{x}^k_i - x^{k+1}_i||}{||\hat{x}^k_i - x^{k+1}_i|| + ||\hat{x}^k_i - x^{k-1}_i||}
\label{eq:eq_feature_04}
\end{eqnarray}
The relative location of the averaged neighbor points can be successfully encoded based on the presented angle- and distance-based descriptions. Figure \ref{fig:kernel_01} illustrates the entire process of feature extraction, {\color{black}and figure \ref{fig:candidates} illustrates candidates of the neighbor points corresponding to extracted features. By using our features, we can represent the relative positions of neighbors accurately.}

\begin{figure*}[t]
\begin{center}
\includegraphics[width=0.9\linewidth]{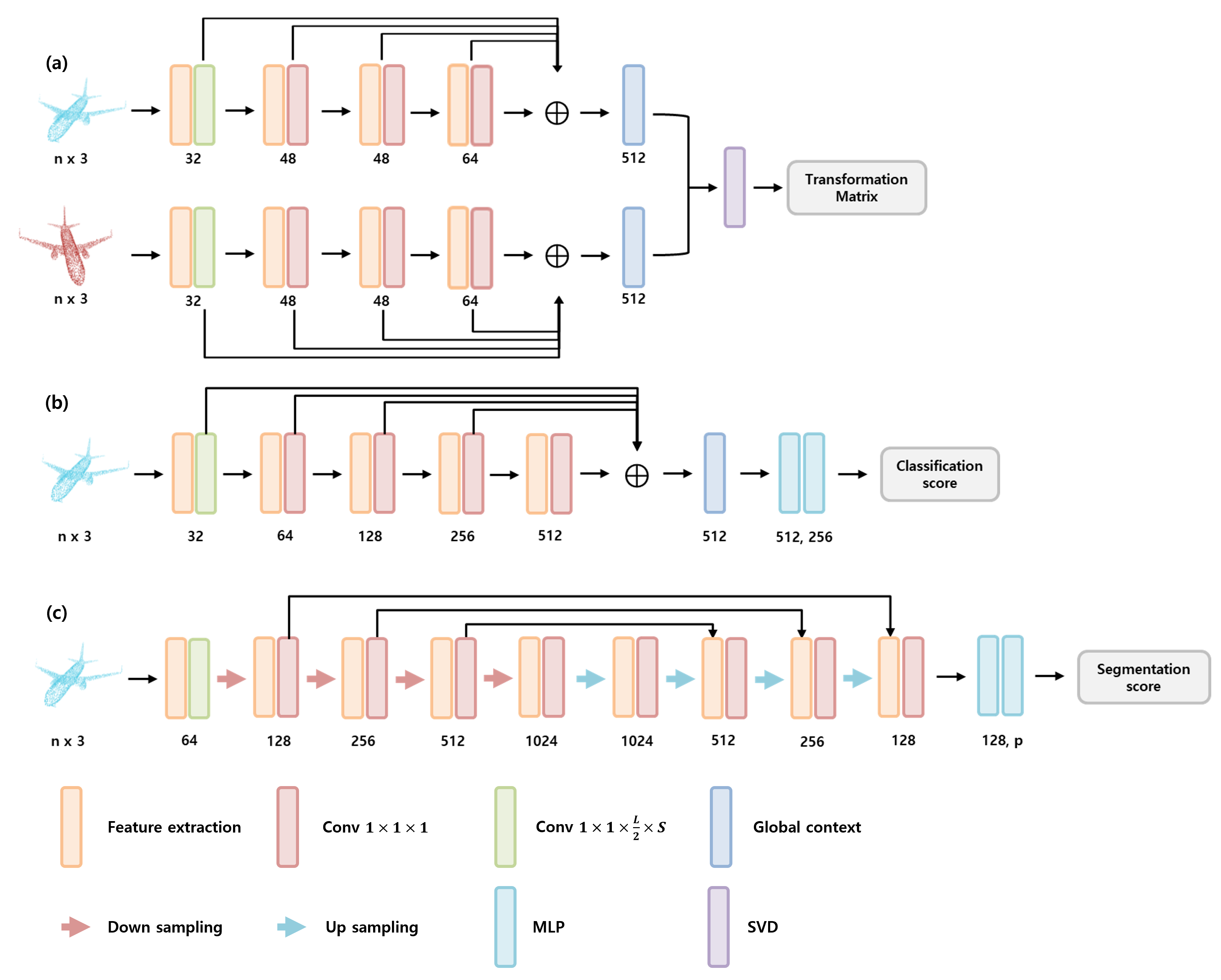}
\end{center}
   \caption{{\color{black} Network architectures for (a) registration, (b) classification, and (c) part segmentation.}}
\label{fig:arc}
\end{figure*}

\subsection{Circular convolution}
Because the kernels are not aligned to the unique local reference frames, the order of the kernels may change depending on the point distribution. However, the adjacent kernels within the cylinder layer are invariant to rotation. Therefore, we extended 1 × 1 × 1 channel-wise convolution based on (\ref{eq:eq_conv_01}) to (\ref{eq:eq_conv_02}):

\begin{eqnarray}
x_i = \sum_{\hat{x}^k_i} f(g(\hat{x}^k_i) ),
\label{eq:eq_conv_01}
\end{eqnarray}

\begin{eqnarray}
x_i= \sum_{\hat{x}^k_i} f(g(\hat{x}^{c(k, -1)}_i), g(\hat{x}^k_i), g(\hat{x}^{c(k, +1)}_i)),
\label{eq:eq_conv_02}
\end{eqnarray}
where $c(k, +1)$ and $c(k, -1)$ indicate the clockwise and counterclockwise adjacent kernels of the $k$-th kernel in the cylinder layer. Subsequently, to avoid the sign problem, kernels are divided into two groups: the collection of kernels above the tangent plane and the collection of kernels below the tangent plane. If the kernel belongs to the first group, we select the clockwise adjacent kernel in a clockwise order. Otherwise, we select the kernel in a counterclockwise order.

In addition, we processed convolution with multiple layers if the layers belong to the same group, i.e., (\ref{eq:eq_conv_02}) is extended to
\begin{eqnarray}
x_i= \sum_{\hat{x}^k_i} f(g(\hat{x}^k_i), g(\hat{x}^{j}_i)_{j \in adj(k)}),
\label{eq:eq_conv_03}
\end{eqnarray}
where $adj(k)$ indicates a set of adjacent kernel points of the $k$-th kernel point in the same group. A circular padding convolution operation was used to implement (\ref{eq:eq_conv_03}). Figure \ref{fig:kernel_03} illustrates the convolution process using the kernels. After convolution, the kernel features around the interest point are aggregated by the summation and maximum value selection.

% \begin{figure}[t]
% \begin{center}
% \includegraphics[width=1.0\linewidth]{img/Fig_scale_arc.png}
% \end{center}
%   \caption{Scale adaptation module. The descriptors are build with the roughly selected kernel sizes. The output is the interpolation weight between two kernel sizes.}
% \label{fig:scale_arc}
% \end{figure}

\subsection{Scale adaptation module}
{\color{black} 
If a target object has an unusual shape when compared to the training data, the normalization process might fail to resolve the scale problem. Because performing normalization cannot resolve the scale problem completely, we first normalized the target objects, and subsequently performed the scale adaptation module on the normalized objects to supplement scale-robustness.
}

To develop a scale-robust descriptor, we adjusted the kernel size based on an analysis of the multiscaled features ($d$ in (\ref{eq:eq_kernel_01})). We first extracted multiple features using multiple kernel sizes (feature extraction in Fig. \ref{fig:overview}). Subsequently, we concatenated the multiscaled features and estimated the interpolation weights between the kernel sizes. Simple convolution operations were employed for the scale analysis, as illustrated in Fig. \ref{fig:overview}. Finally, the output size of the kernel was used to encode the proposed descriptor for CNN encoding (Fig. \ref{fig:overview}).
% Figure \ref{fig:scale_arc} illustrates the scale adaptation module. 

\subsection{CNN encoder architectures}
The designed CNN encoders are illustrated in Fig. \ref{fig:arc}. 
For the registration task, four feature-extraction layers are initially used. Using a shortcut connection, the multiscale features are concatenated. Subsequently, global contexts from the concatenated features are estimated to improve the representations (as described in the following subsection, Section F). Inspired by the deep closest point (DCP) method \cite{DCP}, the singular value decomposition (SVD) module is used to estimate the transformation matrix. 
For the classification and segmentation tasks,  {\color{black} we used the downsampling and upsampling operations which reduces and increase the number of points by using subsampling. 
Subsequently, Additional fully connected layers (multi-layered perceptron) are used to estimate the scores.}

% =======================================================================

\begin{figure}[h]
\begin{center}
\includegraphics[width=1.0\linewidth]{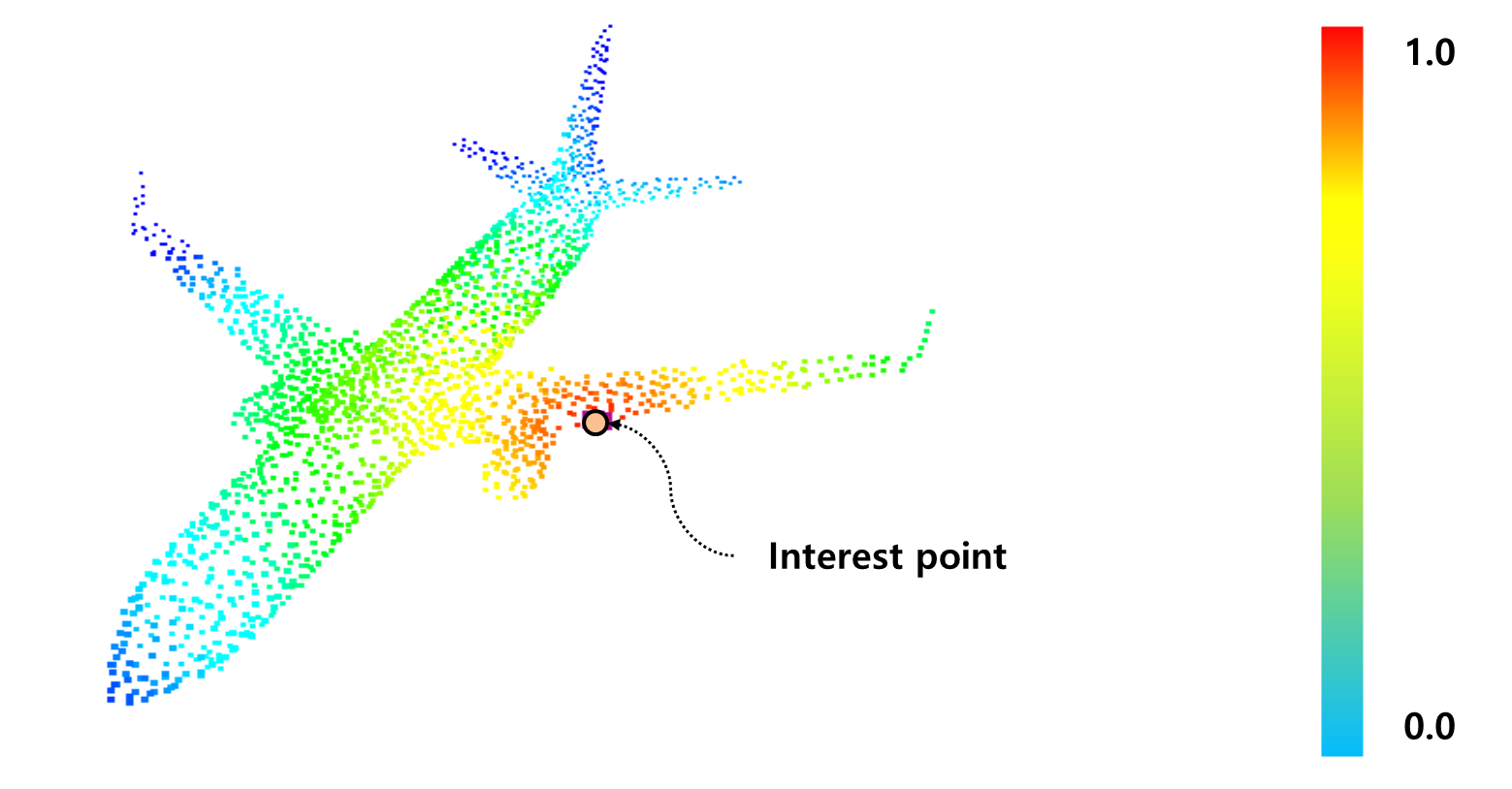}
\end{center}
   \caption{Colors of point cloud indicate global context weights of the interest point. Red indicates a weight value of one and blue indicates a weight value of zero.}
\label{fig:global}
\end{figure}

\subsection{Aggregating global context}
Local descriptors of monotonous and repeating areas are typically considered as nonsalient descriptors. To improve representations of the descriptor, we estimated the global context from local features (global context module in Fig. \ref{fig:arc}). Rather than estimating a single global context for all points using max-pooling, we estimated the adaptive global contexts for each point by using distance-based weights. 

To estimate the global feature for the $i$-th point, weights $w_{ij}$ are calculated based on the Gaussian distance between the $i$-th and $j$-th points (weights for an interest point in Fig. \ref{fig:global}). Subsequently, the averaged local features are estimated using the weights $w_{ij}$ for all $j$. 
\begin{eqnarray}
g_i = \sum \frac{w_{ij} f_j}{\sum w_{ij}} \text{    where    } w_{ij} = exp(\frac{-(x_j - x_i)^2}{d^2}) 
\label{eq:eq_global}
\end{eqnarray}
Once the global contexts are estimated for each point, the global contexts are concatenated to each local feature. Finally, a convolution operation is performed on concatenated features.

\section{Result}
We implemented three tasks: registration, classification, and segmentation. For the registration and classification tasks, we used the ModelNet40 database \cite{ShapeNet}. For the segmentation task, we used the ShapeNetPart database \cite{ShapeNetPart}. For the registration task, we compared our methods with PointNetLK \cite{PointNetLK} and DCP \cite{DCP}. In the case of classification and part-segmentation tasks, we compared our methods with several methods, such as PointNet \cite{PointNet}, DGCNN \cite{DGCNN}, and KPConv \cite{KPConv}.

We analyzed our method on the registration task using the ModelNet40 database \cite{ShapeNet}. ModelNet40 contains 12,311 meshed computer-aided design models from 40 categories. ModelNet40 is split by category into training and test sets; the first 20 categories among the 40 categories were used for training. For each model, 1,024 points were used for training and testing.

The DCP method \cite{DCP} presented an end-to-end network for a rigid registration. Inspired by this method, we used the SVD module to compute a rigid transformation. Figure \ref{fig:arc}(a) illustrates the registration architecture. As evaluation metrics, the mean squared error, root mean squared error, and mean absolute error were used for rotation and translation.

% ===============================================================
\begin{table*}[h]
\begin{center}
\begin{tabular}{|l|c|c|c|c|c|c|c|c|c|}
\hline
Method & R-MSE & R-RMSE & R-MAE & T-MSE & T-RMSE & T-MAE & C \\
\hline\hline
ICP & 892.601 & 29.876 & 23.626 & 0.086 & 0.293 & 0.251 & -\\
Go-ICP \cite{Go-ICP} & 192.258 & 13.865 & 2.914 & 0.000 & 0.022 & 0.006 & -\\
FGR \cite{FGR} & 97.002 & 9.848 & 1.445 & 0.000 & 0.013 & 0.002 & -\\
PointNetLK \cite{PointNetLK} & 306.323 & 17.502 & 5.280 & 0.000 & 0.028 & 0.007 & 20\\
PointNetLK \cite{PointNetLK} & 227.870 & 15.095 & 4.225 & 0.000 & 0.022 & 0.005 & 40\\
DCP-v2 \cite{DCP} & 9.923 & 3.150 & 2.007 & 0.000 & 0.005 & 0.003 & 20 \\
DCP-v2 \cite{DCP} & 1.307 & 1.143 & 0.770 & 0.000 & 0.001 & 0.001 & 40 \\
\hline\hline
Our method & \textbf{0.017} & \textbf{0.130} & \textbf{0.064} & \textbf{0.000} & \textbf{0.000} & \textbf{0.000} & 20\\
\hline
\end{tabular}
\end{center}
\caption{{\color{black}Global registration results for ModelNet40.} The evaluation metrics are mean squared error (MSE), root mean squared error (RMSE), and mean absolute error (MAE) for rotation (R-) and translation (T-). C indicates the number of categories for training.}
\label{table:regi}
\end{table*}
% ===============================================================

% ===============================================================
\begin{table*}[h]
% 1
\begin{center}
\begin{tabular}{|c|c|c|c|c|c|c|}
\hline
Method & R-MSE & R-RMSE & R-MAE & T-MSE & T-RMSE & T-MAE\\
\hline
\hline
%ICP & 1217.618 & 34.894 & 25.455 & 0.086 & 0.293 & 0.251\\
%Go-ICP \cite{Go-ICP} & 157.072 & 12.533 & 2.940 & 0.0009 & 0.031 & 0.010\\
%FGR \cite{FGR} & 98.635 & 9.932 & 1.952 & 0.0014 & 0.038 & 0.007\\
%PointNetLK \cite{PointNetLK} & 526.401 & 22.943 & 9.655 & 0.0037 & 0.061 & 0.033\\
%DCP-v2 \cite{DCP} & 95.431 & 9.769 & 6.954 & 0.0010 & 0.034 & 0.025\\
%PRNet \cite{PRNET} & 24.857 & 4.986 & 2.329 & 0.0004 & 0.021 & 0.015\\
%SpinNet \cite{spinnet} & - & - & - & - & - & -\\
ICP & 297.080 & 17.236 & 8.610  & 0.007 &  0.082 & 0.043 \\
Go-ICP \cite{Go-ICP}  & 184.199 & 13.572 & 3.416  &  0.002 &  0.045 & 0.015 \\
FGR \cite{FGR}  & 40.832 &  6.390 & 1.240  & 0.001 &  0.038 & 0.008 \\
PointNetLK \cite{PointNetLK}  & 334.67 & 18.294 & 9.730  &  0.008 &  0.092 & 0.053 \\
DCP \cite{DCP}  & 45.617 & 6.754 & 4.366  & 0.004 &  0.061 & 0.040 \\
PRNet \cite{PRNET} & 7.355 &  2.712 & 1.372  & 0.000 &  0.017 & 0.012 \\
FMR \cite{FMR} & 25.412 & 5.041 & 2.304 & 0.001 & 0.038 & 0.016\\
IDAM \cite{IDAM} & 46.950 & 6.852 & 1.761 & 0.003 & 0.054 & 0.014\\
DeepGMR \cite{DeepGMR} & 356.832 & 18.890 & 9.322 & 0.008 & 0.087 & 0.056\\
OMNet \cite{OMNet}  & 4.322 & 2.079 & 0.619 & 0.000 & 0.018 & 0.008\\
SpinNet \cite{spinnet} & 1.355 & 1.164 & 0.902 & 0.000 & 0.013 & 0.011\\
\hline
\hline
%Our method & 1.330 & 1.153 & 0.784 & 0.0005 & 0.023 & 0.015\\
Our method & \textbf{0.741} & \textbf{0.861} & \textbf{0.440} & \textbf{0.000} & \textbf{0.008} & \textbf{0.004}\\
\hline
\end{tabular}
\end{center}
\caption{{\color{black} Partial registration results for ModelNet40.}}
\label{table:partial_reg}
\end{table*}
% ===============================================================

\begin{figure*}[h]
\begin{center}
\includegraphics[width=0.85\linewidth]{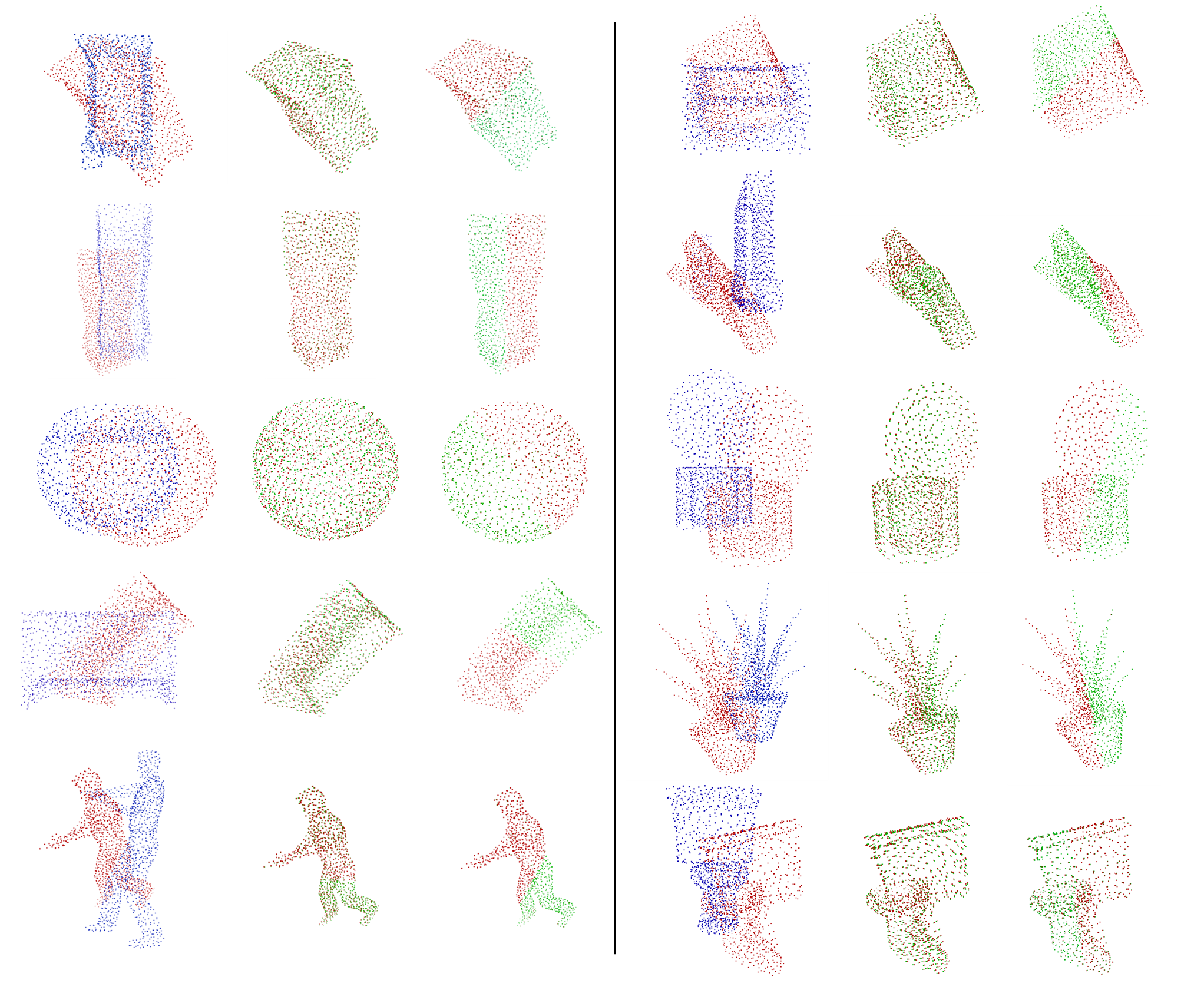}
\end{center}
   \caption{Left: source (blue) and target (red) point clouds, Middle: registration results of deep closest point method \cite{DCP}. Green indicates the transformed source point clouds. Right: registration results of our method. }
\label{fig:result_regi}
\end{figure*}

% ========================================
\begin{table}[tbp]
\begin{center}
\begin{tabular}{|l|c|c|c|c|}
\hline
Method & OA & mIoU & mcIoU & RI\\
\hline\hline
SPLATNet \cite{SPLATNet} & - & 85.4 & 83.7 & X\\
SGPN \cite{SGPN} & - & 85.8 & 82.8 & X\\
3DmFV-Net \cite{3DmFV} & 91.6 & 84.3 & 81.0 & X\\
SynSpecCNN \cite{SynSpecCNN} & - & 84.7 & 82.0 & X\\
RSNet \cite{RSNet} & - & 84.9 & 81.4 & X\\
SpecGCN \cite{SpecGCN} & 91.5 & 85.4 & - & X\\
PointNet \cite{PointNet} & 89.2 & 83.7 & 80.3 & X\\
PointNet++ \cite{PointNet++} & 90.7 & 85.1 & 81.9 & X\\
KD-Net \cite{Kd-net} & 90.6 & 82.3 & 77.4 & X\\
SO-Net \cite{SO-Net} & 90.9 & 84.9 & 81.0 & X\\
PCNN by Ext \cite{PCNN} & 92.3 & 85.1 & 81.8 & X\\
SpiderCNN \cite{SpiderCNN} & 90.5 & 85.3 & 82.4 & X\\
MCConv \cite{MCConv} & 90.9 & 85.9 & - & X\\
FlexConv \cite{FlexConv} & 90.2 & 85.0 & 84.7 & X\\
PointCNN \cite{PointCNN} & 92.2 & 86.1 & 84.6 & X\\
DGCNN \cite{DGCNN} & 92.2 & 85.2 & 85.0 & X\\
SubSparseCNN \cite{3DmFV} & - & 86.0 & 83.3 & X\\
KPConv \cite{KPConv} & 92.9 & 86.2 & \textbf{85.1} & X\\
ShellNet \cite{ShellNet} & 93.1 & - & - & X\\ 
Point Transformer \cite{transformer} & 93.7 & \textbf{86.6} & 83.7 & X\\ 
PAConv \cite{paconv}  & \textbf{93.9} & 86.1 & 84.6 & X\\ 
\hline
RI-ShellConv \cite{RI-ShellConv} & 86.5 & 80.3 & 75.3 & O\\ 
ClusterNet \cite{Clusternet} & 87.1 & - & -& O\\ 
REQNN \cite{REQNN} & 83.0  & - & -& O\\ 
PRIN \cite{PRIN} & - & 71.1 & 67.6 & O\\ 
RIF \cite{TVCG2021} & 89.4 & 82.5 & 79.4 & O\\
\hline
Our method & \textbf{93.2} & \textbf{85.9} & \textbf{83.3} & O\\
\hline
\end{tabular}
\end{center}
\caption{{\color{black} ModelNet40 classification results (i.e. OA) and ShapeNetPart segmentation results (i.e. mIoU). OA, mIoU, and mcIoU indicate the overall accuracy, instance average intersection over union, and class average intersection over union, respectively. RI indicates whether the method is the rotation-robust method or not.}}
\label{table:clsseg}
\end{table}
% ========================================

\subsection{Registration}

As listed in Table \ref{table:regi}, our method significantly reduced the registration errors when compared to the other methods {\color{black} since the methods used translation-invariant features, not rotation-invariant features.} Even when compared to the results trained with all categories, our method showed better performance.

{\color{black} We conducted the partial registration task by using the sampled point clouds from ModelNet40 inspired by PRNet \cite{PRNET}. The overall architecture is similar to the registration architecture. One difference is that we computed a rigid transformation using a RANSAC \cite{RANSAC} with the generated descriptors. Table \ref{table:partial_reg} lists the partial registration results. 
SpinNet \cite{spinnet} aligned and mapped the point patch to the cylindrical volume to capture detailed geometric information. Subsequently, the method used the continuous convolution method to capture the geometric structure in a rotation-invariance manner. 
However, there are three drawbacks: 1) Severe memory consumption, 2) the optimal patch size, and 3) the sign problem. On the contrary, our proposed method consumed relatively a small amount of computational memory when compared to SpinNet \cite{spinnet} and automatically determined the kernel size using the scale adaptation module. Moreover, our method concerned the sign problem by using the symmetric circular convolution method. 
As a result, our method significantly reduced the registration errors with simple architecture and less computation memory.
}

Figure \ref{fig:result_regi} illustrates the registration results of the DCP \cite{DCP} and proposed methods. The results of the DCP method \cite{DCP} showed a small error between the two point clouds. Conversely, the results of our method showed superior matching performance. These results indicate that the proposed descriptor matches the feature-based correspondences between the source and target points in a superior manner when compared to the DCP \cite{DCP}.
% =======================================================================

% ==============================================================
\begin{table*}[b]
% 1
\begin{center}
\begin{tabular}{|c|c|c|c|c|c|c|c|c|}
\hline
Conv method & R-MSE & R-RMSE & R-MAE & T-MSE & T-RMSE & T-MAE \\
\hline
channel-wise & 0.040420 & 0.201046 & 0.105576 & 0.000000 & 0.000149 & 0.000094  \\
\hline
circular conv & \textbf{0.017159} & \textbf{0.130991} & \textbf{0.064475} & \textbf{0.000000} & \textbf{0.000048} & \textbf{0.000027} \\
\hline
\hline

%3
\hline
KNN & R-MSE & R-RMSE & R-MAE & T-MSE & T-RMSE & T-MAE \\
\hline
10 & 0.017159 & 0.130991 & 0.064475 & 0.000000 & 0.000048 & 0.000027 \\
\hline
2 & \textbf{0.014517} & \textbf{0.120486} & \textbf{0.065029} & \textbf{0.000000} & \textbf{0.000037} & \textbf{0.000025} \\
\hline
\hline

%4
\hline
Global information & R-MSE & R-RMSE & R-MAE & T-MSE & T-RMSE & T-MAE \\
\hline
X & 0.017159 & 0.130991 & 0.064475 & 0.000000 & 0.000048 & 0.000027  \\
\hline
O & \textbf{0.008142} & \textbf{0.091766} & \textbf{0.046526} & \textbf{0.000000} & \textbf{0.000047} & \textbf{0.000027}  \\
\hline

\end{tabular}
\end{center}
\caption{Parameter and ablation study for ModelNet40 global registration task.}
\label{table:study}
\end{table*}
% ===============================================================

% ==============================================================
\begin{table*}[h]
% 1
\begin{center}
\begin{tabular}{|c|c|c|c|c|c|c|c|c|}
\hline
Scale adaptation (training/test scale) & R-MSE & R-RMSE & R-MAE & T-MSE & T-RMSE & T-MAE\\
\hline
X (1.00/0.50) & 0.111988 & 0.334647 & 0.112045 & 0.000000 & 0.000044 & 0.000031 \\
\hline
X (1.00/1.00) & 0.017159 & 0.130991 & 0.064475 & 0.000000 & 0.000048 & 0.000027 \\
\hline
X (1.00/1.50) & 0.023768 & 0.154168 & 0.080950 & 0.000000 & 0.000150 & 0.000082 \\
\hline
\textbf{X (Total)} & \makecell{0.050971 $\pm$ \\ 0.043229} 
& \makecell{ 0.206601 $\pm$ \\ 0.0910345 }  
& \makecell{ 0.085823 $\pm$ \\ 0.019723 } 
& \makecell{ 0.000000 $\pm$ \\ 0.000000 }  
& \makecell{ 0.000080 $\pm$ \\ 0.000049 } 
& \makecell{ 0.000046 $\pm$ \\ 0.000025 }  \\
\hline
O (1.00/0.50) & 0.041671 & 0.204134 & 0.073383 & 0.000000 & 0.000056 & 0.000036 \\
\hline
O (1.00/1.00) & 0.021910 & 0.148021 & 0.055841 & 0.000000 & 0.000039 & 0.000025 \\
\hline
O (1.00/1.50) & 0.027835 & 0.166837 & 0.064757 & 0.000000 & 0.000073 & 0.000046 \\
\hline
\textbf{O (Total)} & \makecell{ 0.030472 $\pm$ \\ 0.008280 } 
& \makecell{ 0.172997 $\pm$ \\ 0.023318 }  
& \makecell{ 0.064660 $\pm$ \\ 0.007161 } 
& \makecell{ 0.000000 $\pm$ \\ 0.000000 }  
& \makecell{ 0.000056 $\pm$ \\ 0.000013 } 
& \makecell{ 0.000035 $\pm$ \\ 0.000008 }  \\
\hline

\end{tabular}
\end{center}
\caption{{\color{black} ModelNet40 global registration results with different scales. (1.0/$\#$) is the experiments in which networks are trained with the original scale of point clouds and tested with ($\#$) scale of point clouds. Values indicate the mean and standard deviation of the results.}}
\label{table:regScale}
\end{table*}
% ===============================================================

% ========================================
\begin{table}[h]
\begin{center}
\begin{tabular}{|l|c|c|c|}
\hline
Method & OA (ZR/AR) & OA (AR/AR) & RI\\
\hline\hline
PointNet \cite{PointNet} & 16.4 & 75.5 & X\\
PointNet++ \cite{PointNet++} & 28.6 & 85.0 & X\\
PointCNN \cite{PointCNN} & 41.2 & 84.5 & X\\
DGCNN \cite{DGCNN} & 20.6 & 81.1 & X\\
ShellNet \cite{ShellNet} & 19.9 & 87.8 & X\\ 
KPConv \cite{KPConv} & 47.8 & 87.8 & X\\
\hline
RI-ShellConv \cite{RI-ShellConv} & 86.5 & 86.5 & O\\ 
ClusterNet \cite{Clusternet} & 87.1 & 87.1 & O\\ 
REQNN \cite{REQNN} & 83.0 & - & O\\ 
RIF \cite{TVCG2021} & \textbf{89.4} & 89.3 & O\\
\hline
Our method & 89.0 & \textbf{91.0} & O\\
\hline
\end{tabular}
\end{center}
\caption{{\color{black} Classification results with rotations for ModelNet40. ZR and AR indicate the azimuthal rotations (ZR) and arbitrary rotations (AR), respectively.}}
\label{table:clsRotation}
\end{table}
% ========================================

% ========================================
\begin{table}[h]
\begin{center}
\begin{tabular}{|l|c|c|c|}
\hline
Method & mcIoU (ZR/AR) & mcIoU (AR/AR) & RI\\
\hline\hline
PointNet \cite{PointNet} & 37.8 & 74.4 & X \\
PointNet++ \cite{PointNet++} & 48.3 & 76.7 & X\\
PointCNN \cite{PointCNN} & 34.7 & 71.4 & X\\
DGCNN \cite{DGCNN} & 37.4 & 73.3 & X\\
ShellNet \cite{ShellNet} & 47.2 & 77.1 & X\\ 
KPConv \cite{KPConv} & 46.3 & 75.8 & X\\
\hline
RI-ShellConv \cite{RI-ShellConv} & 75.3 & 75.3 & O\\ 
PRIN \cite{PRIN} & 64.6 & 67.6 & O\\ 
RIF \cite{TVCG2021} & \textbf{79.2} & 79.4 & O\\
\hline
Our method  & 73.7 & \textbf{80.1} & O\\
\hline
\end{tabular}
\end{center}
\caption{{\color{black} Results of ShapeNetPart segmentation. mcIoU indicates the class average intersection over union. The second and third column (i.e., ZR/AR and AR/AR) indicate values of mcIoU obtained from training the data based on ZR and AR, respectively, and tested on AR.}}
\label{table:clssegARAR}
\end{table}
% ========================================

% =======================================================================

\subsection{Classification and Segmentation}
We analyzed the classification and part-segmentation performances of our method using ModelNet40 \cite{ShapeNet} and ShapeNetPart \cite{ShapeNetPart} databases, respectively. ModelNet40 contains 12,311 models. Among the models, 9,843 models were used for training, and the remaining 2,468 models were used for testing. For each model, 1,024 points were used for training and testing. ShapeNetPart contains 16,681 models from 16 categories. Each point is annotated using part labels. For each model, we used 2,048 points for training and testing. 

Figures \ref{fig:arc}(b) and (c) illustrate the classification and part-segmentation architectures, respectively. 
Table \ref{table:clsseg} lists the classification and part-segmentation results. While considering the evaluation metrics, the overall accuracy was used for Modelnet40 classification and mean intersection over union was used for ShapeNetPart segmentation. 
{\color{black} 
We compared our methods with the non-rotation-invariant methods and rotation-invariant methods.
The non-rotation-invariant methods typically represented the relationship between points based on point coordinates.
Unlike the non-rotation-invariant methods, the rotation-invariant methods used rotation-invariant features (e.g. relative distance and angle) to achieve rotation-invariant property.
However, it is hard to represent the accurate relationship between points with rotation-invariant features, and convolution with more than one point in rotation-invariant-order is a challenging problem.
Therefore, as listed in Table \ref{table:clsseg}, the non-rotation-invariant methods showed better performances when compared to the rotation-invariant methods under non-rotation environments.

To address the problems, our proposed method represented neighbors with not only the rotation-invariant features, but the kernels to supplement the relationship representation. Further, we used the circular convolution method, which processed the adjacent kernels simultaneously to capture the relationships between the points.
Therefore, as clearly demonstrated in Table \ref{table:clsseg}, our proposed descriptor outperformed the rotation-invariant methods and achieved comparable performance when compared to the non-rotation-invariant methods. 
Because the non-rotation-invariant methods typically used point coordinates as the features, it was easy to learn geometric information based on each point location.
On the contrary, to develop a rotation-robust descriptor, we used rotation-invariant features. Moreover, we aligned our kernels to the normal vector, and it means that the order of the kernels may change depending on the point distribution unlike the non-rotation-invariant methods. 
These properties affected the method performance, and thus our method showed inferior performances when compared to the state-of-the-art non-rotation-invariant methods under non-rotation environments.
However, rotation-invariance is a desired feature for real-world applications. Thus, it is significant that our proposed method achieved superior accuracy among the rotation-invariant methods.
}

% =======================================================================
\begin{comment}
\begin{figure}[t]
\begin{center}
\includegraphics[width=1.0\linewidth]{img/Fig_result_seg.png}
\end{center}
   \caption{Left: source points; Right: part-segmentation results}
\label{fig:result_seg}
\end{figure}
\end{comment}
% =======================================================================

\begin{figure*}[t]
\begin{center}
\includegraphics[width=0.8\linewidth]{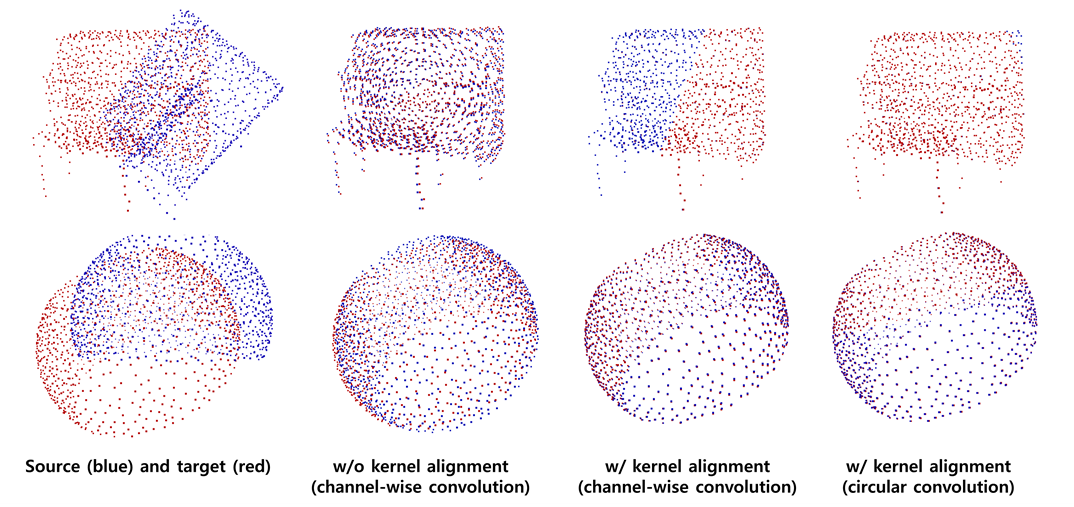}
\end{center}
   \caption{Comparison of methods}
\label{fig:comparison}
\end{figure*}

% =======================================================================
\subsection{Parameter and ablation study}

We conducted several parameter and ablation studies on the registration task to verify the effect of our method (Table \ref{table:study}), i.e., by varying the following parameters: convolution operation, the number of nearest neighbors for each kernel, and global context.

First, we experimented with the convolution methods, i.e., 1 × 1 × 1 channel-wise convolution and circular convolution methods. The network with circular convolution significantly improved the performance in terms of both rotation and translation. These results indicate that the circular convolution operations successfully captured the geometric features based on adjacent kernels. Figure \ref{fig:comparison} illustrates the registration results according to the kernel alignment and convolution methods. As illustrated, the network with the aligned kernel-based circular convolution showed better registration results.

Second, we analyzed our method with a different number of neighbors. As listed in Table \ref{table:study}, the errors were not significantly dependent on the number of neighbors. Because we used the distance-based weights for each neighbor, the closer neighbors had more influence. Owing to the use of the distance-based weights, employing the averaged neighbors reduced the influence of the number of neighbors.

Third, we conducted registration with the global context. Consequently, the rotation error decreased significantly when compared to the other experiments. These results demonstrate that the global context resolves the ambiguities of each local descriptor.

{\color{black} 
\subsection{Robustness study}
In addition to the parameter studies on the registration task, we conducted scale- and rotation- robustness studies for evaluation. First, we experimented with the scale adaptation module. The network is trained with the original scale (1.00) of point clouds and tested with different scales (0.50, 1.50) to demonstrate the scale robustness. Table \ref{table:regScale} lists the results of different scale tests for each model. The mean and standard deviations are presented, and the proposed network showed stable results when the network used the scale adaptation module. The results indicate that the scale adaptation module determined the optimal kernel size to capture geometric information so that the performance of the global registration outperformed.\par

In addition, to demonstrate the rotation robustness, we trained the network with azimuthal rotations (around the gravity axis) (ZR) and arbitrary rotations (AR) and tested with arbitrary rotations (ZR/AR, AR/AR) for the classification and segmentation (Table \ref{table:clsRotation} and \ref{table:clssegARAR}). (-/-) indicates which rotational metric was used for training/testing, respectively.\par

We compared the results with state-of-the-art rotation-invariant methods to demonstrate the rotation robustness. The rotation-invariant methods \cite{RI-ShellConv}\cite{Clusternet} used rotation-invariant features to represent the relative positions of neighbors and processed each point using MLP to avoid processing in non-rotation-invariant order.
However, since there is no additional reference point, the used rotation-invariant features were insufficient to completely represent the relative positions \cite{RI-ShellConv}\cite{Clusternet}. RIF \cite{TVCG2021} used the additional reference points, but the reference points have a chance to be changed depending on object shape variation, and it may result in insufficient consistency of the descriptors between similar object parts. Moreover, the shared MLP simply processed each point feature without considering other points.\par

On the contrary, by using the kernels (i.e., reference points) which have fixed distances from an interest point, our proposed method can represent the relative positions of neighbors accurately. Moreover, by using the circular convolution method which processed the adjacent kernels at once to capture the relationships between the information, our descriptor improved geometric information representations.\par

As a result, in the case of training and evaluating under arbitrary rotation (AR/AR), our method showed superior performances when compared to the state-of-art methods.
In addition, in the case of training under azimuthal rotation and evaluating under arbitrary rotation (ZR/AR), the accuracy losses of our methods were not significant when compared to the other non-rotation-invariant methods since our method used rotation-invariant features.
}

\section{Discussion}
Point cloud analysis requires rotation- and scale-robust feature representation. {\color{black} It is challenging to develop a robust descriptor with a good benchmark accuracy}. In this paper, we propose an aligned kernel-based feature representation to resolve these limitations. To make the descriptor robust to rotation, we aligned the kernels to the local reference frame. Subsequently, we applied normal sign-independent convolutions to the descriptors rather than using fixed kernels that are independent of the rotations \cite{KPConv}. Instead of using translation-invariant features \cite{DGCNN}, we used rotation-robust features from the aligned kernels. In addition, to improve the representations of the descriptor, we estimated the adaptive global context for each point rather than using a single global context \cite{PPFNet}. Because the kernel-based descriptors are highly dependent on the size of the given kernels, we adjusted the kernel size based on the trainable weights. 
\par

The experimental results for various tasks (i.e., registration, classification, and part segmentation) showed promising results for feature representation. In the registration task, the rotation and translation errors decreased significantly. This indicates that our descriptors successfully captured the salient and corresponding geometric information between the two transformed point clouds. {\color{black} In the case of classification and segmentation tasks, our proposed method showed the best performances under rotations.} These results indicate that our method is not only limited to the transformed point cloud task but also applicable to general purposes (i.e., feature representations). Several parameter and ablation studies have also demonstrated that our proposed methods improved the feature representations and stability of the descriptor.

\section{Conclusion} 
Encoding rotation- and scale-robust features is a challenging task for point cloud representation. The robustness of each parameter is critical for a successful application in various downstream tasks. In this paper, we proposed a CNN-based feature encoding method to resolve this task. The proposed kernel alignment, feature projection, and kernel-conscious convolution methods demonstrated superior performance on the registration task when compared to previous methods. Moreover, the proposed scale adaptation and global aggregation methods successfully captured the optimum scale parameter and global geometric features for each local descriptor, respectively.

% =======================================================================
%-------------------------------------------------------------------------

\ifCLASSOPTIONcaptionsoff
  \newpage
\fi

\bibliographystyle{IEEEtran}
\bibliography{main.bib}

\end{document}